\documentclass[10pt, conference, compsocconf]{IEEEtran}
\usepackage{graphicx}
\usepackage{float}
\usepackage{multirow}

\ifCLASSOPTIONcompsoc
  \usepackage[nocompress]{cite}
\else
  \usepackage{cite}
\fi

\ifCLASSINFOpdf
 
\else
 
\fi

\usepackage{color}
\usepackage[table,xcdraw]{xcolor}

\usepackage{hyperref}
\hypersetup{
	bookmarks=false,
	colorlinks=true,       % false: boxed links; true: colored links
	linkcolor=blue,          % color of internal links (change box color with linkbordercolor)
	citecolor=green,        % color of links to bibliography
	filecolor=magenta,      % color of file links
	urlcolor=cyan         
}

\makeatletter
\newcommand{\newlineauthors}{%
  \end{@IEEEauthorhalign}\hfill\mbox{}\par
  \mbox{}\hfill\begin{@IEEEauthorhalign}
}
\makeatother

\newcounter{todocounter}

\usepackage{amsmath}
\usepackage{amssymb}

\makeatletter
\newcommand{\linebreakand}{%
  \end{@IEEEauthorhalign}
  \hfill\mbox{}\par
  \mbox{}\hfill\begin{@IEEEauthorhalign}
}
\makeatother

\title{A title}
\author{ 
\IEEEauthorblockN{Tuan Nguyen}
\IEEEauthorblockA{\textit{Faculty of Mathematical Economics} \\
\textit{National Economics University}\\
Hanoi, Vietnam \\
nttuan@neu.edu.vn}
\and
\IEEEauthorblockN{Phong Nguyen\IEEEauthorrefmark{2}}
\IEEEauthorblockA{\textit{Department of Advanced Interdisciplinary Studies} \\
\textit{The University of Tokyo}\\
Tokyo, Japan \\
xphongvn@gmail.com}
\linebreakand

\IEEEauthorblockN{Hanh Pham, Truong Bui, Tan Nguyen, Duc Luong}
    \IEEEauthorblockA{School of Information and Communication Technology}
    \IEEEauthorblockA{Hanoi University of Science and Technology} 
    \IEEEauthorblockA{Hanoi, Vietnam}
    \IEEEauthorblockA{\{hanhpv.183525, truong.bm183646, tan.np183824, duc.lh183712\}@sis.hust.edu.vn}

}
\begin{document}

\title{SP-GPT2: Semantics Improvement in Vietnamese Poetry Generation}

\maketitle
\begingroup\renewcommand\thefootnote{\IEEEauthorrefmark{1}}
\footnotetext{Authors contribute equally}
\begingroup\renewcommand\thefootnote{\IEEEauthorrefmark{2}}
\footnotetext{Corresponding author}
\begin{abstract}
Automatic text generation has garnered growing attention in recent years as an essential step towards computer creativity. Generative Pre-training Transformer 2 (GPT2) is one of the state-of-the-art approaches that have excellent successes. In this paper, we took the first step to investigate the power of GPT2 in traditional Vietnamese poetry generation. In the earlier time, our experiment with base GPT2 was quite good at generating the poem in the proper template. Though it can learn the patterns, including rhyme and tone rules, from the training data, like almost all other text generation approaches, the poems generated still has a topic drift and semantic inconsistency. To improve the cohesion within the poems, we proposed a new model SP-GPT2 (semantic poem GPT2) which was built on the top GPT2 model and an additional loss to constrain context throughout the entire poem. %First, our model can learn sentence embedding, in which the abstract context representation from the poetry of each pair of two sentences can be learned. Then we constrained the sentence embedding of each pair in the poem is similar. Thus, it is guaranteed that the meaning of the poem is consistent. 
For better evaluation, we examined the methods by both automatic quantitative evaluation and human evaluation. Both automatic and human evaluation demonstrated that our approach can generate poems that have better cohesion without losing the quality due to additional loss. %Determining a grading technique for Luc-Bat, which is traditional Vietnamese poetry and has complex rules, was one of the difficulties we encountered while researching this issue. 
At the same time, we are the pioneers of this topic. We released the first computational scoring module for poems generated in the template containing the style's rule dictionary. %Since there was no  Luc-Bat dataset, we collected, cleaned, and created our training data,
Additionally, we are the first to publish a Luc-Bat dataset, including 87609 Luc Bat poems, which is equivalent to about 2.6 million sentences, combined with about 83579 poems in other styles was also published for further exploration.  The code is available at https://github.com/fsoft-ailab/Poem-Generator
\end{abstract}

\IEEEpeerreviewmaketitle

%===========================================================
\begin{figure}[bt]
    \centerline{\includegraphics[width=1\linewidth]{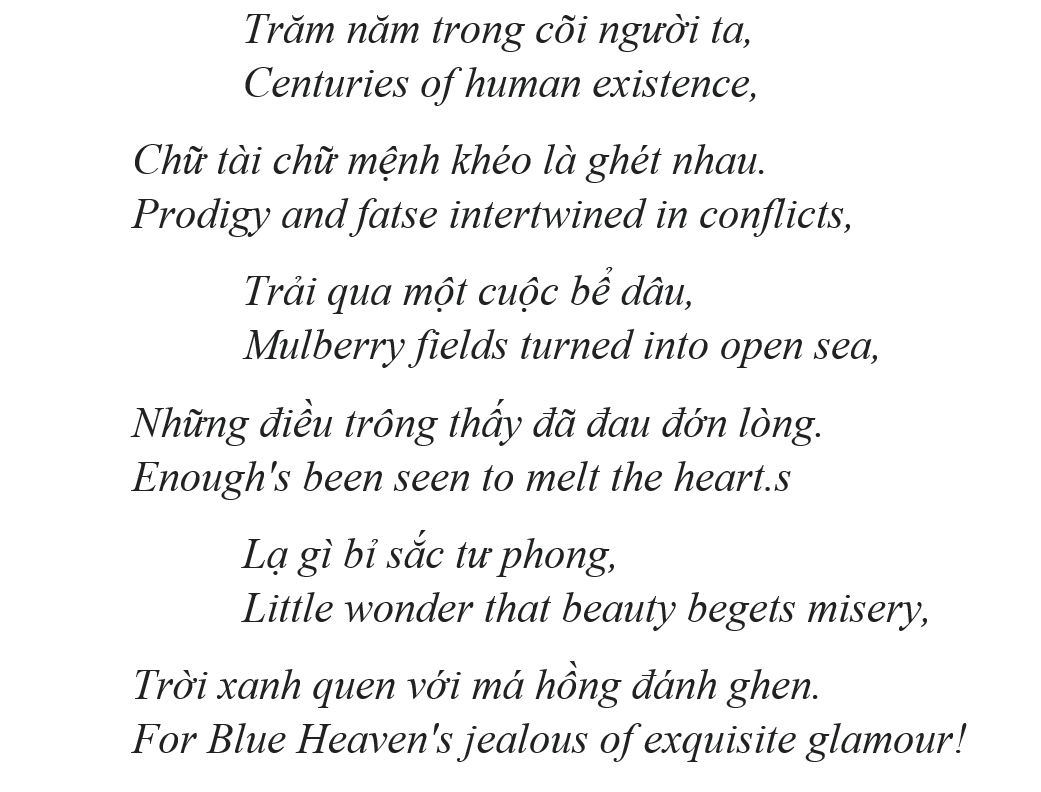}}
    \caption{A sample of Luc-Bat poem}
    \label{fig:sample}
    \vspace{-5pt}
 \end{figure}
\section{Introduction}
Luc Bat is an ancient style of Vietnamese poetry, translated as "six-eight"- the number of syllables in each line.  
Although there is no predetermined length,  each line must alternate between six and eight syllables, as the rhyming system is relatively complicated. Not only the rhyme, but its tone and flow are also intricate.  Nonetheless,  this form of poetry also follows an explicit template in form.  
Besides the context represented throughout the poem,  each stanza is quite abstract with many literary implications such as embodiments,  metaphors and comparisons.  Fig.  \ref{fig:sample} shown above is an instance.  Despite its complexity,  it adheres to a few basic principles:
\begin{itemize}
\item The 6\textsuperscript{th} word's rhyme in the first line is relevant to the 6\textsuperscript{th} word's rhyme in the second line. 
\item The 6\textsuperscript{th} word's rhyme in the third line and the 6\textsuperscript{th} word's rhyme in the fourth line is relevant to the 8\textsuperscript{th} word's rhyme in the second line. 
\item Tones in each 6-word sentence are level,  oblique and level in the 2\textsuperscript{nd},  4\textsuperscript{th} and 6\textsuperscript{th} word's tone respectively. 
\item Tones in each 8-word sentence are level,  oblique,  level and level corresponding to the 2\textsuperscript{nd},  4\textsuperscript{th},  6\textsuperscript{th} and 8\textsuperscript{th}  word's tone. 
\end{itemize}

Natural language processing (NLP)  is a discipline of computer science,  specifically the branch of artificial intelligence (AI).  It enables computers to interpret texts and spoken words in the same manner as humans.  NLP blends computational linguistics (human language rule-based modeling) with statistical,  machine learning,  and deep learning models.  These technologies provide computers the ability to analyze human language in the form of text or speech to understand its full meaning,  including speaker's or writer's purposes and sentiments. 

% Researches involving natural language processing have recently attracted great attention recently,  as we can witness from the bursting work on natural-language generation.  Natural-language generation aims to generate sentences that perform some purposes,  including poem generation production in human languages.  We reached a higher level on tackling more cognitive work related to Vietnamese poetry generation,  which needs to be in the right form of rhyme and tone.  Our paper will introduce some methods to work on some tasks: generation of poetic language and how to control output’s context in our case. 

In the natural language processing field,  related poem generation problems have been studied,  the authors mainly focused on the quality of styles and rhythms.  There were also some publications related to text generation, or poem generation.  For a style with templates,  the traditional method was to generate poems using rhyme indexing based on sample sentences, which are concatenated afterwards.  This is very good at the form,  but the lingual outputs are all from prior knowledge that does not sound interesting.  Moreover,  the context in this generation approach is tough to be relevant in each sentence.  As a result,  the consistency context in continuous sentences has poor performance.  In the industrial field,  Facebook has proposed to generate English rhythmic poetry with neural networks \cite{hopkins-kiela-2017-automatically},  and Microsoft has developed a system called XiaoIce \cite{xiaoice},  in which poem generation is one of the most essential features.  Nevertheless, generating poems with context control remains a new topic with grand challenges. 

Compared with typical types of document generation like the magazine or other essential text generation, which only focuses on generating a descriptive sentence with no rules,  generating a good poem with context control is a more challenging problem.  There is a broader difference between textual descriptions and poetic symbols inspired by visualization and imagination.  For instance, the symbol of "afternoon" found in a poem might be used to describe further sorrows and the loneliness of a woman waiting for her husband to return from the battlefield.  
% Furthermore,  there was not any Vietnamese Poetry dataset published before this paper for our research,  this causes us a lot of trouble in the early stages of development. 

In this paper,  we introduced the SP-GPT2 model to constrain the meaning throughout the poems generated.  With a standard GPT2 \cite{radford2019language} approach,  the model can learn the patterns, including Luc Bat rhyme and tone template based on training data patterns.  This approach gave us an excellent result in template check, an average 84.54 score, with our evaluation module described in \ref{autoeval}.  But there was still a problem with the context representation of the output.  So we proposed the SP-GPT2 model that can keep the context throughout the poem.  We computed the punishment for the GPT2 model.  Every two next-to-sentence embeddings,  in which the abstract context represents poetry,  can be learned pair by pair. 
% and then updating weights via policy-gradient method has shown significant effectiveness to our solution.  
% \textbf{Such a framework assures that major poetic information important for poem production may be recognized and modeled from such extended pairings.  }
Our method has shown significant outperformance by reaching 86.94 average scores in the template and achieving a score of 3.34 in the context assessed by the specialized while the pure GPT2 gets only 3.02. 

To evaluate our generated poems,  we first need to figure out how rhymes and tones in each Luc Bat stanza works.  They both are complex. The rules of Luc Bat poems have been mentioned in many documents, but there were no scoring and evaluating modules.  So we took a further step to build a module for scoring each poem generated by rhymes and tones.  
%Each verse has so many relatives because Vietnamese grammar has all five tones classified into two types in Luc-Bat: level and oblique.

Overall,  the contributions of the paper are as follows:
\begin{itemize}
    \item We presented SP-GPT2, which can generate Luc Bat poem while controlling for context representation throughout the output,  which plays an important role in boosting poetic aesthetics. 
    \item We proposed the first computational evaluation module for template check,  which took a lot of effort to build a dictionary of riles. 
    \item We published the first huge dataset in the Vietnamese poetry area,  including Luc Bat and other genres. 
\end{itemize}

The rest of the paper is structured as follows.  First,  Section \ref{relatedwork} introduces related work.  Then,  Section \ref{models} describes the details of the proposed model. Section \ref{resultandeval} presents the experimental setup and evaluation results. Finally, a conclusion of the paper and discussion on future work are illustrated in Section \ref{conclusion}.

%===========================================================
\section{Related Work} \label{relatedwork}
Automatic natural language generation, especially poetry generation, has been an attractive topic for researchers in decades.  One of the first computational implementations that go beyond mere mechanical creativity often relied on rule-based or template-based reasoning methods, for example, rule-based \cite{10.1007/978-1-4471-0275-5_2} ,
phrase search  \cite{10.1007/978-3-540-89222-9_26} \cite{10.1007/978-3-642-04052-8_19} template search \cite{Oliveira2012PoeTryMeA}, genetic search \cite{genericalgo}, text summarization \cite{yan-rui-jiang}.  Besides, various statistical machine translation methods were implemented \cite{jiang-zhou-2008-generating} \cite{he-jing-zhou}.  However, these approaches have the same problem: they are based on the surface forms of words or characters, with no deep comprehension of the content of the poem. 

For a few years, with the powerful ability of natural language representation in neural models \cite{bengio-ducharme} \cite{goldberg-yoav}, different models have taken full use of and shown significant progress.  
% As this is the first paper for the Vietnamese poetry generation, the language has many similarities in structure with Chinese, we only review studies on the Chinese poetry generation. 
The first approach of generating fluent poems, vanilla recurrent neural networks \cite{rnn}, was found in the works \cite{zhang-lapata-2014-chinese} \cite{hopkins-kiela-2017-automatically}. But with this RNN approach, we could not avoid theme drifts caused by this long-sequence generation.  Reference \cite{wang-qixin-2016} utilized the attention mechanism \cite{bahdanau}  by which human intention and all the characters that have been generated so far guide the progress for quatrain generation.  
% Text generation, including poem generation, is commonly viewed as a sequence-to-sequence issue, in which the encoder-decoder framework is used to encode the preceding sequence and the decoder produces the later sequence (Wang, Luo, and Wang 2016; Yi, Li, and Sun 2017).   

All the neural models mentioned above attempt to generate fluent and meaningful poems with rules in rhymes and tones, which can be learned from corpus patterns but there were no considerations on improving contexts throughout the poem generated.  Most recently, researchers have suggested many new methods of poetry generation, such as employ memory networks \cite{wang-can-machine} \cite{yi-generating-rnn}, Variational AutoEncoder (VAE) \cite{yang-gen-vae} and reinforcement learning \cite{yi-etal-2018-automatic}.  Related to our work, a transformer-based language model, GPT2 \cite{radford2019language} transformer-based language model is proven to be good at natural language generation, which performs very well on casual texts with context controlling.  However, for abstract theme of each poem, which is a typical of the Vietnamese poetry, GPT2 shows less power in terms of theme and semantic relevance.  These approaches have no semantic refinements. 

Despite significant improvements in the poetry generation, the abstract context through the Vietnamese poem generated is still a problem.  To improve semantics, we refined the model output in each pair of sentences generated to constraint the context by embedding with the previous sentence.  Besides, our model is self-supervised and can be trained well with unlabeled data. 
\section{Models}\label{models}
\subsection{GPT2}
GPT2\cite{radford2019language} decoder is based on Transformer architecture for text generation. After each token is produced, it is added to a sequence of inputs. Then the new sequence becomes the input to the model in next step. The idea is called “auto-regression”. In self-attention calculation, future tokens are masked, which blocks information from tokens that are located to the right of the position being calculated.
\begin{figure}[htbp]
\centerline{\includegraphics[width= 10cm]{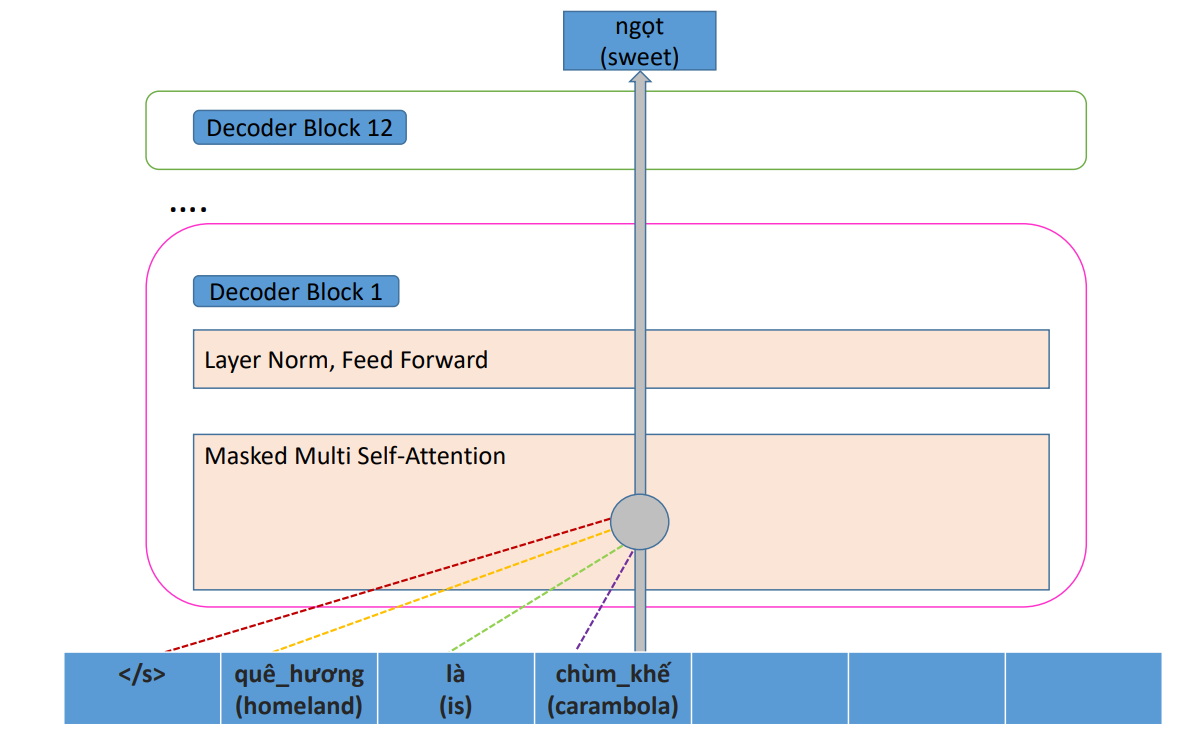}}
\caption{Decoder block}
\label{fig}
\end{figure}
Output of GPT2 is calculated as follows:
\begin{equation}
   \displaystyle h_t = \gamma(\sum_{i=1}^{t-1} w_i*h_i)
\end{equation}
\begin{equation}
   \displaystyle G_t = \mbox{softmax}(w_{t}*h_{t} )
\end{equation}
\begin{itemize}
  \item $h_{t}$ is the output of t-th token of final hidden layer.
  \item $w_1, . . . , w_{t}$ are trainable parameters.
  \item $G_t$ is the GPT2 output of t-th token.
\end{itemize}
GPT2 \cite{radford2019language} was trained with a causal language modeling (CLM) task and is therefore powerful at predicting the next token in a sequence. The final hidden vectors corresponding to the present token are fed into a fully connected layer, then a softmax function to produce the probability vector with the same size as the vocabulary. The causal language modeling objective is a cross-entropy loss on predicting the next token. 
In our experiment with GPT2, at first, we used BPE (Byte Pair Encoding)\cite{neural-machine-translation} to build our tokenizer, which training on 2.6M sentences. Then, we trained GPT2 \cite{radford2019language} with 12 layers, 768 hidden units, and 12 heads. The model was optimized with Adam\cite{adam} using the following parameters $\beta_1 = 0.9$, $\beta_2 = 0.999$, $\epsilon = 1\mathrm{e}{-6}$ and L2 weight decay of 0.01.
\subsection{SP-GPT2}
\subsubsection{RNN, LSTM}
A recurrent neural network, also known as RNN\cite{10.5555/65669.104451} is a memory network having hidden states. RNN takes into account historical information by showing the dependency of the current token with previous tokens. However, RNN has the difficulty of accessing information for a long time ago. This is because, during backpropagation, RNN suffers from the vanishing gradient problem, which occurs when the gradient shrinks as it back propagates through time. To deal with this problem, LSTM \cite{lstm} was introduced to provide not only short-term memory but also long-term memory.
\begin{figure}[htbp]
\centerline{\includegraphics[width= 10cm]{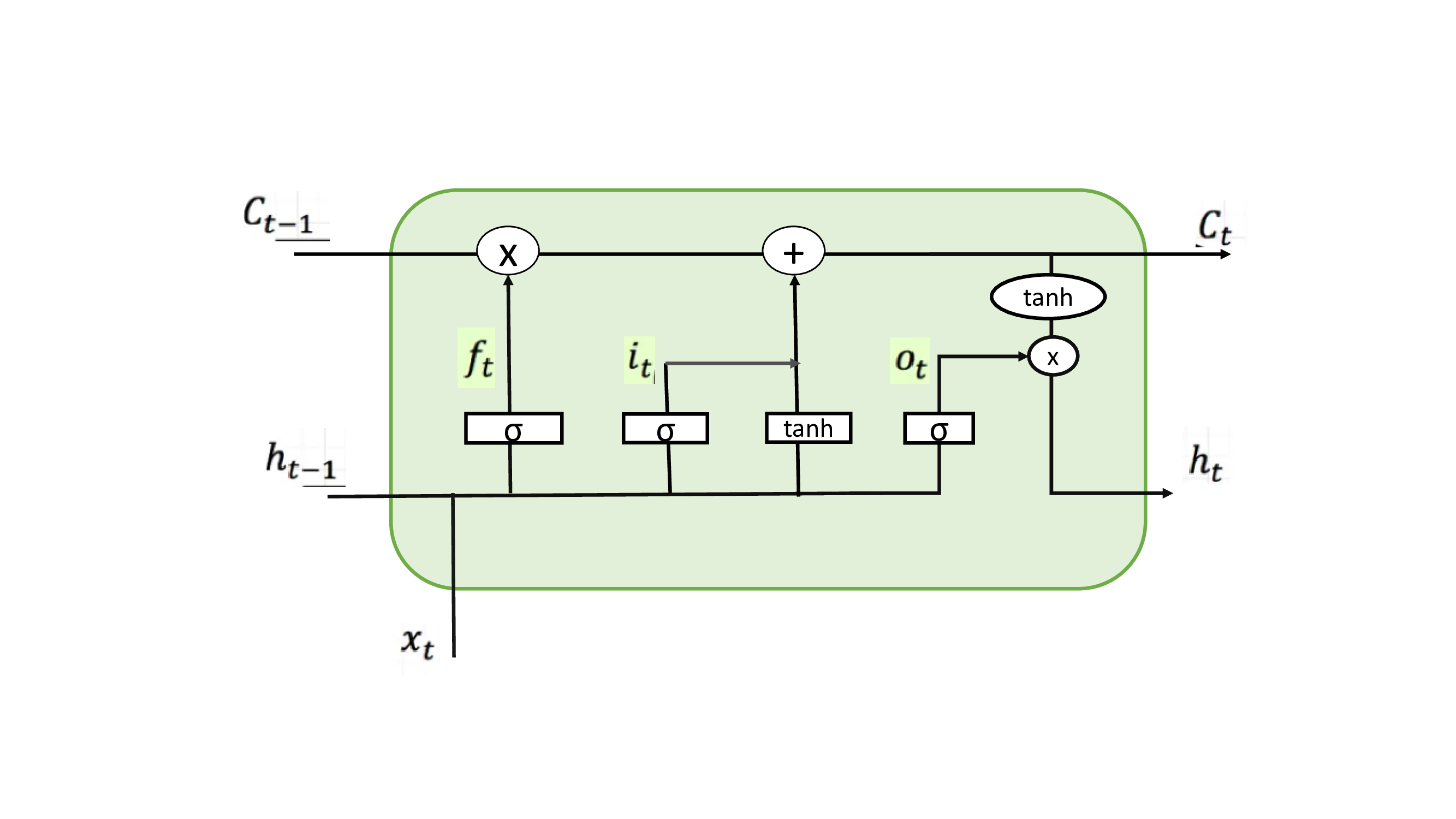}}
\caption{Illustrated LSTM model}
\label{fig}
\end{figure}
\begin{itemize}
  \item Forget gate: $\displaystyle f_t = \sigma(U_f*x_t + W_f*h_{t-1} + b_f)$
  \item Input gate: $\displaystyle i_t = \sigma(U_i*x_t + W_i*h_{t-1} + b_i)$
  \item Output gate: $\displaystyle o_t = \sigma(U_o*x_t + W_o*h_{t-1} + b_o)$
\end{itemize}
with c is cell state and h is hidden state.

Output of LSTM contains: $c_t$, $h_t$. Input of LSTM contains: $c_{t-1}$, $h_{t-1}$, $x_t$ is input at time t; $c_{t-1}$, $h_{t-1}$ is output of previous layer; $b_f, b_i, b_o$ are bias coefficients; $W, U$ are learned model parameters.
\subsubsection{Contextual vector}
\begin{figure}[htbp]
\centerline{\includegraphics[width= 8cm]{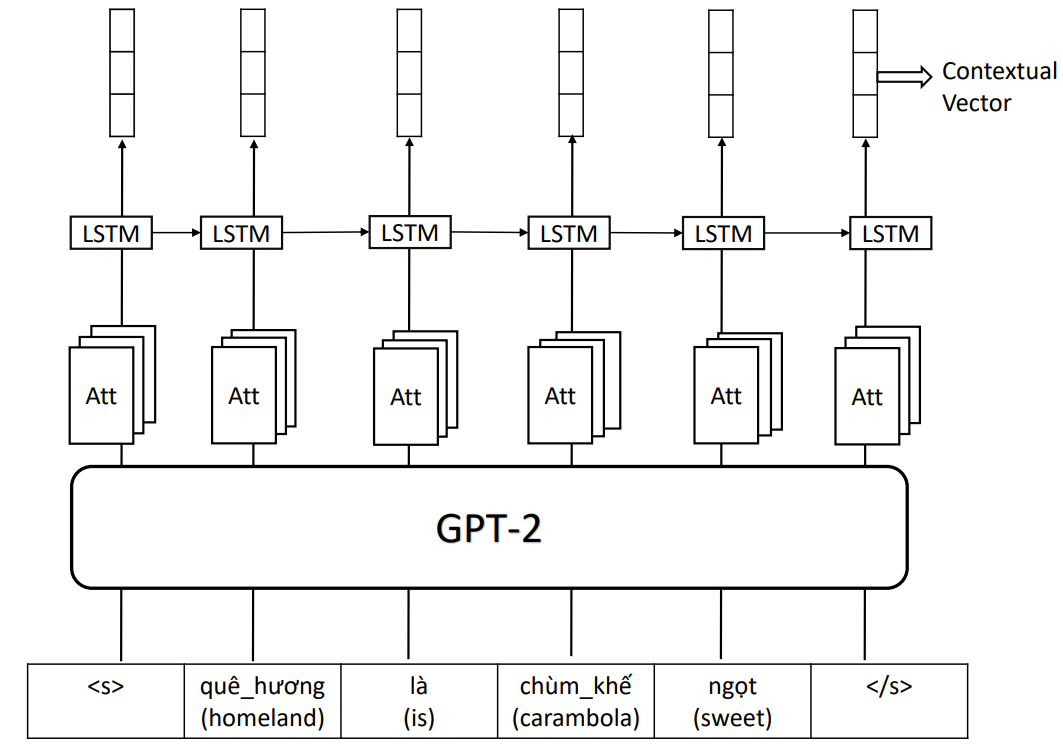}}
\caption{Contexual vector model}
\label{fig}
\end{figure}
Contextual vector is a meaning representation of the input sequence. After the input sequence went through the GPT2 model, we got a sequence of high level embedding of each token. We then passed these embedding through the self-attention layer. The self-attention mechanism allows the embedding to interact with each other and figure out who they should pay more attention to. The outputs are aggregates of these interactions and attention scores. Then, we used the LSTM layer to capture long term dependency of the output sequence. Finally, the last hidden state was chosen as contexual vector of the whole input.
\subsubsection{Custom loss}
Luc Bat is a poetic form that needs ensuring standards in terms of grammar and meaning. During the training with GPT2, we recognized that GPT2 was only grammatically learned and that meaning was not consistent throughout the stanza. Particularly, Luc Bat has rhymes in two consecutive sentences in a stanza, so two pairs of sentences 6-8 in the stanza must have the same contextual in the whole stanza. To help the model understand the meaning of the poem, we designed a two-part loss function to drive the similarities between pairs of sentences in the poem. The first term is the categorical cross-entropy, a common loss function, and minimizing it increases
the probability of generating the ground truth word from the softmax output. The second term
is a mean squared error loss to calculate the loss between two vectors representing the content of two verses in the stanza. This loss can improve semantic binding in a stanza. Intuitively, the model learns to control contextual poems by minimizing MSE loss. The custom loss is shown in Fig. \ref{fig5}. These were our training steps:
\begin{enumerate}
  \item We first trained with the categorical
  cross-entropy to help model generate format of Luc-Bat
  \begin{equation}
      loss = \frac{1}{M-1} \sum_{i=1}^{M-1} -log(\frac{exp(L[\textnormal{k}])}{\sum_{j=1}^{V}exp(L[j])})
  \end{equation}
  \item Then we split a stanza into two pairs, each pair contains one six verse and eight verses, and fed them to our model to get embedding
\begin{equation} 
\begin{split}   
       customloss & = \frac{1}{M-1} \sum_{i=1}^{M-1} -log(\frac{exp(L[\textnormal{k}])}{\sum_{j=1}^{V}exp(L[j])})\\
       & +\sum_{i=1}^{N}(E_{prev} - E_{next})^2
\end{split}
\end{equation}
\end{enumerate}
\begin{itemize}
  \item V: vocab size of tokenizer
  \item M: number of token in one block
  \item L: prediction scores of the language modeling head (scores for each vocabulary token).
  \item k: id of next token.
  \item $E_{prev}, E_{next}$ : contextual vector of two pairs 6-8 verse in one stanza poem.
  \item N: the number of pairs of verse 6-8 in a block.
\end{itemize}
\begin{figure}[htbp]
\centerline{\includegraphics[width= 9.5cm]{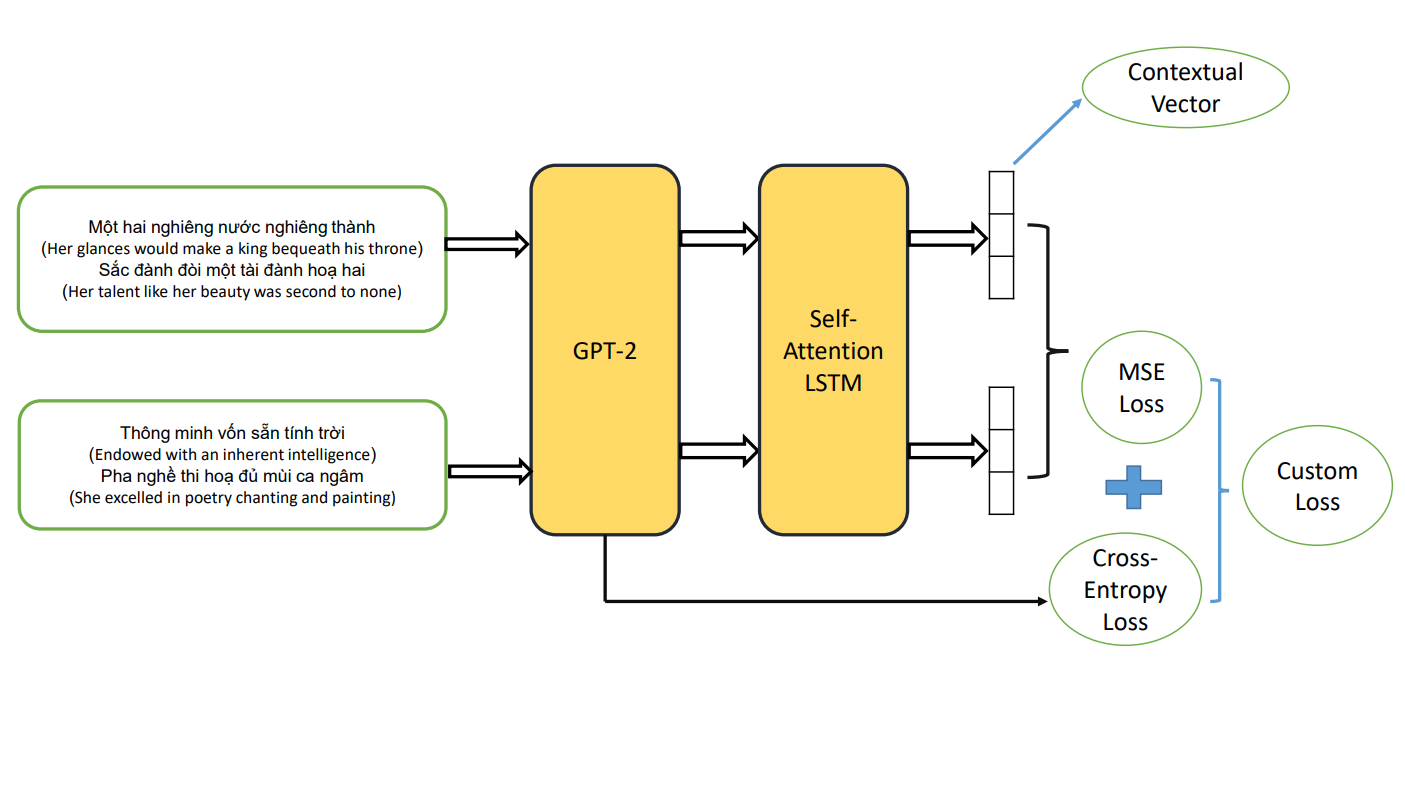}}
\caption{SP-GPT2 model}
\label{fig5}
\end{figure}
%===========================================================
\section{Result and evaluation} \label{resultandeval}
\subsection{Dataset}
As Luc Bat is a Vietnamese traditional poetry style that appeared 90 percent in the form of folk by word of mouth, it took us a lot of time to collect all resources that are existing to publish the first widespread public dataset in this art area.  We collected the dataset from many resources that may come from groups, members of groups, who are pretty specialized in Luc Bat recitation, and from other reliable sources. Then all collected poems were checked by our evaluation method introduced in \ref{autoeval} to filter out qualified poems which follow the style rule. We are now confident to release this dataset with high quality and diversity in topics. 

In this paper, we used a large Vietnamese poetry corpus in many genres in which the Luc Bat style has an outstanding number.  Our baseline model is self-supervised learning, so the corpus are only sets of cleaned poems.  Only for Luc Bat, there were more than 87,609 traditional  Vietnamese poems that are collected from reliable resources.  About 2.6 million sentences were found on many topics, e.g., love, family, nature, and even poems about the pandemic COVID-19.  Each poem was split intro quatrains, a stanza with four sentences, and after all, were randomly shuffled to ensure the baseline model could unintentionally learn the poem's patterns.  
% This large quantity of data enables the semantic content of most Vietnamese characters to be dependable.  

All this data was pre-processed then evaluated by our method, with the average score for all quatrains in the dataset is 98.01 out of 100.  After using the whole dataset for training to generate computational poems, with each approach, we sampled a lot of outputs for each model to assess.  Both automatic evaluation and professional human evaluation were utilized. 

\subsection{Automatic Evaluation} \label{autoeval}
Almost every form of poetry has a blueprint, including classical Luc Bat, which sounds impossible owing to its sophisticated rhymes and tones.  It was also one of the most challenging tasks, even though several publications have shown the style patterns and guidelines.  We gathered credible materials to produce the first Vietnamese template assessment module. 

To create this module, we must first create a dictionary that includes all rhymes and tones connected while adhering to some clear criteria derived from reliable sources of information.  

Using the Luc Bat principles, we can readily determine that there are (3*n-1) words that require rhyme checking and (7*n) words with tones in each stanza, where n is the number of sentence pairings.  So we create a scoring module in which the rhyme rules account for 70 percentages of the overall score of a quatrain and tone accounts for 30 percentages.  And the maximum score each stanza can reach is 100.  The scoring formula:
\begin{equation}
   \displaystyle Score = 100 * (1 - \frac{R}{3n-1} - \frac{T}{7*n})
\end{equation}
\begin{itemize}
    \item $R$ is the number of words with wrong rhyme.
    \item $T$ is the number of words with wrong tone.
\end{itemize}
Specifically, if the score is high, the generated poems follow the template rules and vice versa.
\subsection{Compared Methods}
To choose the optimal base for our computational poets, we trained various language models:
\begin{itemize}
    \item \textbf{SP-GPT2}: GPT2 with a custom loss.  
    \item \textbf{GPT2-SL}: GPT2 with syllable level. 
    \item \textbf{GPT2-WL}: GPT2 with word level. 
\end{itemize}

We assessed each model's ability to create formulaic poetry that ensures rhyme and tone requirements by evaluating sets of 500 sampled quatrains.  Each output generated is scored by the module introduced above. 
\begin{figure}[htbp]
\centerline{\includegraphics[width=80mm]{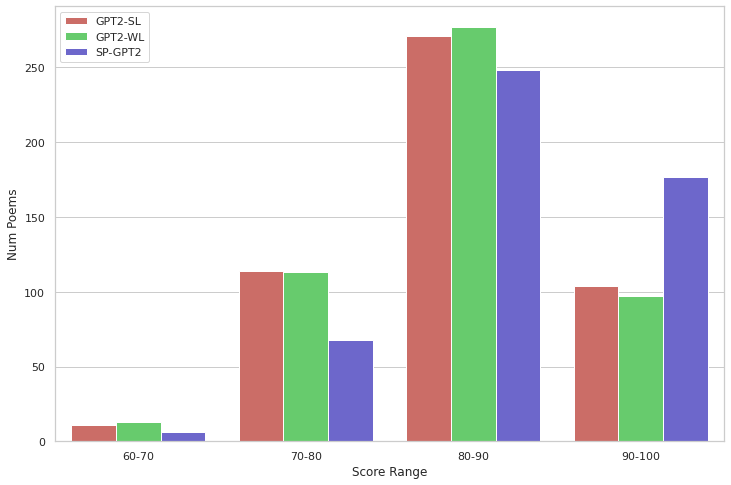}}
\caption{Comparison between models' scores evaluated by automatic module. }
\label{fig2}
\label{fig}
\end{figure}

The Fig. \ref{fig2} illustrates the frequency of scores on three models.  We can see that the total rating of the generated poems was more than 60, it means all three models generated well in the template rules.  With the highest range from 90 to 100, our \textbf{SP-GPT2} model outnumbers the others by about 175 to approximately 100. 
Poems generated by our model emerged in the fewest numbers in the smaller range.  Average score for three models is  86.94, 84.54 and 84.26 for \textbf{SP-GPT}, \textbf{GPT2-SL}, \textbf{GPT2-WL} respectively.  Overall, our model outperformed previous techniques with an additional loss each pair of phrases. 

\subsection{Creativity Evaluation}
While training so many epochs or lacking of data, the model tended to be overfitting.  So we proposed a method to evaluate the creativity of the Vietnamese Poetry Generator. 
We chose randomly 500 context inputs which are the first sentences of data points
in the training set.  We then calculated the ratio of the verses in a generated
poem that appears in the training set to the total number of verses of the generated poem
denoted as $c_{i}$.  We defined a formula below as the creativity score of the poetry creative model: 
\begin{equation}C = \frac{1}{500}\sum_i^{500}(1-c_i) \end{equation}
Specifically, if $C$ value is high, the poetry creative model is highly creative and
vice versa. 
\begin{table}[htbp]
\caption{Creativity Score of Models}
\begin{center}
\begin{tabular}{|c|c|}
\hline
\textbf{Model}&\textbf{Creativity Score} \\
\hline
\textbf{GPT2-SW}   & 0. 955  \\
\hline
\textbf{GPT2-SL}   & 0. 964  \\
\hline
\textbf{SP-GPT2}   & \textbf{0. 970}  \\
\hline
% \multicolumn{2}{l}{$^{\mathrm{a}}$Sample of a Table footnote. }
\end{tabular}
\label{tab1}
\end{center}
\end{table}

Overall, the models we have experimented showed high creativity about generated poems. 
However, our SP-GPT2 model is more creative than the original GPT2 with different level of
language. 
\subsection{Human Evaluation}
\begin{figure}[b]
\centerline{\includegraphics[width=85mm]{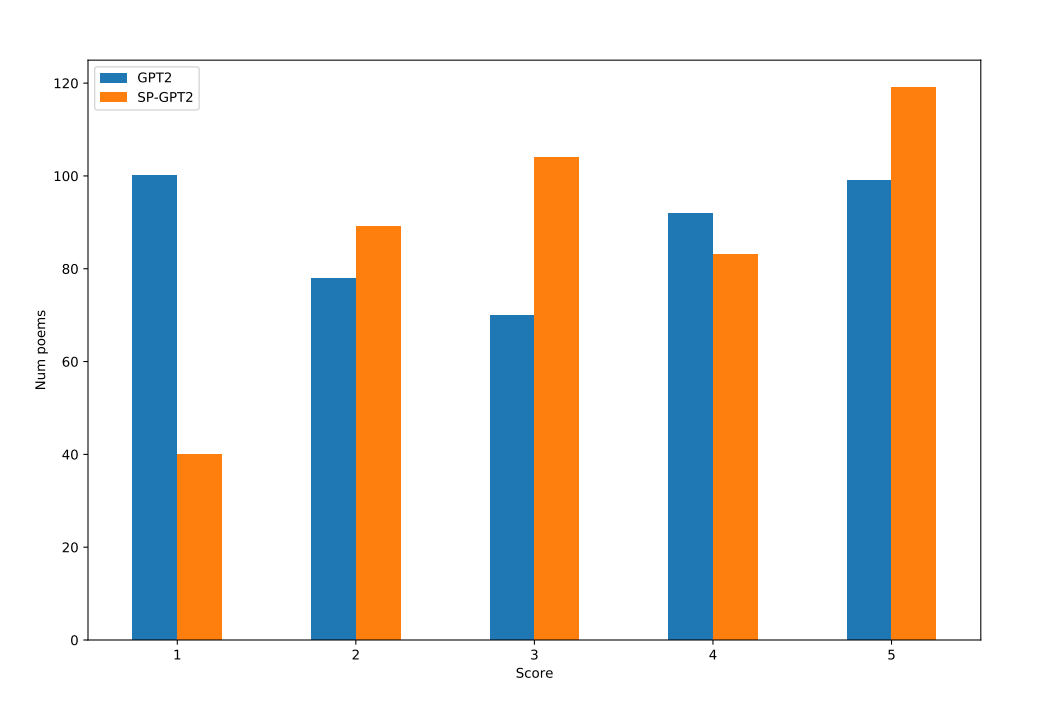}}
\caption{Comparison between models' scores evaluated by a human. }
\label{fig:human_eval}
\end{figure}
Unlike the assessment of poetry rules such as rhymes, metrical law, and six-eight meter, we carried out the human evaluation method in terms of semantic in a poem. 
For the human evaluation method, we chose randomly opening sentences that are abstract from easy to difficult.  Then each GPT2 model and SP-GPT model generated 103 six-eight poems, and we appraised these poems by reading and marking them on a scale from zero to five.  The comparison is shown in Fig.  \ref{fig:human_eval}. 

Regarding the GPT2 model, in terms of semantic, for low-abstract opening verses, the poems generated are relatively good.  However, the semantic relevance between the stanzas is not throughout the entire poem.  For high-abstract opening sentences that we rated as difficult inputs, the performance of GPT2 is not stable, and many poems were confusing to understand in verses. 

 With all the opening sentences, poems generated from our model keep the context of poetry throughout the stanzas.  Poems generated by our model maintained the poetic context throughout the stanzas.  In the higher score range, from three to five, our model outnumbered the best GPT2.  In smaller range, the poems generated by SP-GPT2 appeared to be fewer.  GPT2 had an average semantic score of 3.02, but our SP-GPT2 had a value of 3.34.  Overall SP-GPT2 model outperformed SP-GPT2 while handling high-abstract opening verses. 

%===========================================================
\section{Conclusion} \label{conclusion}
We took another step forward in the field of computational creativity by generating poems with context control. The job entails writing poetry with the following constraints: 1) the poetry should follow a rhyme and tone scheme, and 2) the poem should follow a constraint abstract theme for consistency. We also offered a baseline for the task, which is based on a neural language model pre-trained on our huge self-collected dataset.

The dataset will also be published, with each participant putting in the most effort in terms of collection and cleaning. It includes not just Luc-Bat's poems but also various styles of Vietnamese poetry. This is the first publicly available dataset on Vietnamese poetry.

We performed an automatic evaluation of the generated poems to determine how good the generated pattern is, followed by a professional human evaluation. And we discovered that our SP-GPT model could be good at generating patterns and outperform other automatic approaches by obtaining good ratings. Moreover, our extra constraints contribute to the literary quality of the resulting poetry.

\ifCLASSOPTIONcompsoc
  \section*{Acknowledgments}
\else
  \section*{Acknowledgment}
\fi
We would like to thank FPT Software AI Lab for supporting our
research.

%=========================================================== 
\bibliographystyle{IEEEtran}
\bibliography{references.bib}

\end{document}